\documentclass{article}

\usepackage{microtype}
\usepackage{graphicx}
\usepackage{xcolor}
\usepackage{algorithm}
\usepackage{algorithmic}
\usepackage{subfigure}
\usepackage{booktabs} % for professional tables
\usepackage{mathtools}
\usepackage{comment}
\usepackage{makecell}
\usepackage{afterpage}
\usepackage[normalem]{ulem}

\usepackage{xurl}
\PassOptionsToPackage{hyphens}{url}\usepackage{hyperref}
\usepackage[accepted]{icml2021}
\icmltitlerunning{Continuous Recursive Neural Networks}

\begin{document}

\twocolumn[
\icmltitle{Modeling Hierarchical Structures with Continuous Recursive Neural Networks}




\begin{icmlauthorlist}
\icmlauthor{Jishnu Ray Chowdhury}{to}
\icmlauthor{Cornelia Caragea}{to}
\end{icmlauthorlist}

\icmlaffiliation{to}{Computer Science, University of Illinois at Chicago, United States}

\icmlcorrespondingauthor{Jishnu Ray Chowdhury}{jraych2@uic.edu}
\icmlcorrespondingauthor{Cornelia Caragea}{cornelia@uic.edu}

\icmlkeywords{Machine Learning, ICML}

\vskip 0.3in
]



\printAffiliationsAndNotice{}  % leave blank if no need to mention equal contribution

\begin{abstract}
%Time and again 
Recursive Neural Networks (RvNNs), which compose sequences according to their underlying hierarchical syntactic structure, have performed well in several natural language processing tasks compared to similar models without structural biases. However, traditional RvNNs are incapable of inducing the latent structure in a plain text sequence on their own. Several extensions have been proposed to overcome this limitation. Nevertheless, these extensions tend to rely on surrogate gradients or reinforcement learning at the cost of higher bias or variance. In this work, we propose Continuous Recursive Neural Network (CRvNN) as a backpropagation-friendly alternative to address the aforementioned limitations. This is done by incorporating a continuous relaxation to the induced structure. We demonstrate that CRvNN achieves strong performance in challenging synthetic tasks such as logical inference \cite{bowman-2015-tree} and ListOps \cite{nangia-bowman-2018-listops}. We also show that CRvNN performs comparably or better than prior latent structure models on real-world tasks such as sentiment analysis and natural language inference.\footnote{Our code is available at:\\ \url{https://github.com/JRC1995/Continuous-RvNN}}
\end{abstract}

\section{Introduction}
\label{Introduction}
Constructing a sentence representation is crucial for Natural Language Processing (NLP) tasks such as natural language inference, document retrieval, and text classification. In certain contexts, methods with a bias towards composing sentences according to their underlying syntactic structures \cite{chomsky1957syntactic} have been shown to outperform comparable structure-agnostic methods \cite{socher-etal-2013-recursive, tai-etal-2015-improved, nangia-bowman-2018-listops,choi-2018-learning, maillard_clark_yogatama_2019, havrylov-etal-2019-cooperative, shen-2019-ordered} and some of these structure-aware methods \cite{shen-2019-ordered, liu2020compositional} also exhibit better \textit{systematicity}  \cite{FODOR19883}. Notably, even contemporary Transformer-based methods  \cite{vaswani-etal-2017-attention} have benefited from structural biases in multiple natural language tasks \cite{wang-etal-2019-tree, fei-etal-2020-retrofitting}.

Recursive Neural Networks (RvNNs) \cite{pollack-1990-recursive, socher-etal-2013-recursive} are capable of composing sequences in any arbitrary order as defined by the structure of the input. However, RvNNs follow a hard structure when composing a given sequence. Latent-Tree models such as Gumbel-Tree-LSTM \cite{choi-2018-learning}, despite being capable of adaptively making structural decisions, still follow a hard structure of composition. As such, these models are forced to either make discrete structural decisions or rely on external parsers. In the former case, the model is often forced to rely on surrogate gradients\footnote{We use the term ``surrogate gradients" in the same sense as \citet{martins-etal-2019-latent} to indicate methods that approximate gradients of a discrete argmax-style mapping. Straight-through estimation with Gumbel-Softmax is such an example.} (resulting in increased bias) or reinforcement learning (resulting in increased variance). The increased bias or variance can lead to poor performance \cite{nangia-bowman-2018-listops} unless much care is taken \cite{havrylov-etal-2019-cooperative}. 

In contrast to these RvNN-based methods, chart-based parsers \cite{le-zuidema-2015-forest, maillard_clark_yogatama_2019} can follow a ``soft-structure" by recursively applying weighted average on outputs from multiple possible structures. However, these methods require the construction and maintenance of a chart of vectors, which is comparatively expensive.

Given these challenges, in this work, we propose Continuous Recursive Neural Network (CRvNN), which incorporates a continuous relaxation to RvNNs making it end-to-end differentiable. We aim for the following features:

\begin{enumerate}
    \item \textbf{Automatic task-specific latent tree induction.} Our model is not reliant on ground truth structural labels in any form. 
    \item \textbf{Backpropagation friendly.} This allows our model to be easily integrated as a module within a larger neural architecture and avoid the challenges of reinforcement learning or surrogate gradients \cite{havrylov-etal-2019-cooperative}.
    \item \textbf{Parallelism.} Unlike most prior latent tree models, CRvNN can compose multiple constituents \textit{in parallel} if they are in the same level of hierarchy. This implies that, ideally, CRvNN can recurse over the induced tree-depth, which typically is much shorter than the sequence length. As such, CRvNN can also alleviate difficulties with gradient vanishing or gradient exploding \cite{hochreiter-etal-1991-untersuchungen, bengio-etal-1994-learning}, which can be caused by applying recurrence or recursion over long-length sequences. 
\end{enumerate}

Although chart-based parsers technically share the above features to an extent, they parallelize over multiple paths of composition instead of straightforwardly composing multiple constituents in a single path. Overall, chart-based parsers still need to recurse over the entire sequence length. In contrast, CRvNN can instead halt early based on the induced tree-depth albeit at the cost of more greediness.

\section{Preliminaries}

In this section, we provide a brief introduction to Recursive Neural Networks.

\subsection{Recursive Neural Network (RvNN)}
\label{RvNN}
An RvNN uses some composition function $f$ to recursively compose two given children into their parent:
\begin{equation}
    p = f(x_i, x_{i+1}),
\end{equation}
Here, we consider a sequence $x_{1:n} = (x_1, x_2,\cdots,x_n)$, and a composition function $f$ that takes two child vectors of dimension $d$ as input and returns a parent vector of the same  dimension $d$. The recursion terminates when it composes every node into the root parent. As an example, consider the sequence $x_{1:6}$ 
with an underlying hierarchical structure expressed as: 
$$((x_1, (x_2, x_3)), (x_4, (x_5, x_6))).$$
This kind of structured sequence can be also expressed in terms of a tree. The RvNN operates on the input as:
\begin{equation}
    p = f(f(x_1, f(x_2, x_3)), f(x_4, f(x_5, x_6))).
\end{equation}

\section{Our Approach}
In this section, we describe the technical details of CRvNN. However, before getting into the exact details, we first construct a new but equivalent reformulation of RvNN. This will make it easier to connect CRvNN to RvNN. 
\subsection{Recursive Neural Network: Reformulation}
\label{reformulation}
We now present the aforementioned reformulation of RvNN. Below we describe different components of our reformulated RvNN individually.

\subsubsection{Composition function} 
\vspace{2mm}

The notion of composition function is the same as in \S\ref{RvNN}. 

\setlength{\parskip}{0.5em}
\noindent {\bf Example.} Given a composition function (or recursive cell) $f(r_1, r_2)$, if one input ($r_1$) represents $x_1$, and another input ($r_2$) represents $(x_2, x_3)$, then the output parent will represent $(x_1, (x_2, x_3))$.

\vspace{2mm}
\subsubsection{Existential probability}
\vspace{2mm}

In an RvNN, once two child representations are composed into a parent, the child representations themselves will no longer be needed. The unneeded child representations can be treated as ``non-existent". Based on this idea, we introduce the notion of ``existential probability."

\setlength{\parskip}{0.5em}
\noindent {\bf Definition 1.} We define the notion of an ``existential probability" $e_i$ to denote whether a representation $r_i$ requires further processing ($e_i=1$) or it should be treated as ``non-existing" and hence ignored ($e_i=0$). Every position has an existential probability, which initially is $1$ for all positions. For now, we only consider binary values $e_i \in \{0,1\}$.

\setlength{\parskip}{0.5em}
\noindent {\bf Example.} 
If $r_i$ from position $i$ is composed with $r_j$ from position $j$, then we may update position $j$ with the composed parent representation of $(r_i, r_j)$ and set the existential probability of position $i$ as $0$. Typically, we can just remove the unnecessary representations. However, in this formulation, we keep them with an existential probability. This is a key change that enables us to bring a continuous relaxation to RvNNs as we will discuss later. 

\begin{table*}[t]
\small
\centering
\def\arraystretch{1.2}
\begin{tabular}{ | l | r | r | r | r | r | r | r } 
%\toprule
\hline
\textbf{Positions:} & $\mathbf{1}$ & $\mathbf{2}$ & $\mathbf{3}$ & $\mathbf{4}$ & $\mathbf{5}$ & $\mathbf{6}$\\
\hline
\multicolumn{7}{l}{Iteration 1}\\
%\midrule
\hline
Sequence: & $x_1$ & $\mathbf{x_2}\rightarrow$ & $x_3$ & $x_4$ & $\mathbf{x_5}\rightarrow$ & $x_6$\\
Existential Probability: & 1 & 1 & 1 & 1 & 1 & 1 \\
Composition Probability: & 0 & 1 & 0 & 0 & 1 & 0 \\
\hline
\multicolumn{7}{l}{Iteration 2}\\
\hline
Sequence: & $\mathbf{x_1}\rightarrow$ & \sout{$x_2$} & $(x_2,x_3)$ & $\mathbf{x_4}\rightarrow$ & \sout{$x_5$} & $(x_5, x_6)$\\
Existential Probability: & 1 & 0 & 1 & 1 & 0 & 1\\
Composition Probability: & 1 & $\emptyset$ & 0 & 1 & $\emptyset$ & 0\\
\hline
\multicolumn{7}{l}{Iteration 3}\\
\hline
Sequence: & \sout{$x_1$} & \sout{$x_2$} & $\mathbf{(x_1, (x_2,x_3))}\rightarrow$ & \sout{$x_4$}& \sout{$x_5$} & $(x_4, (x_5,x_6))$\\
Existential Probability: & 0 & 0 & 1 & 0 & 0 & 1 \\
Composition Probability: & $\emptyset$ & $\emptyset$ & 1 & $\emptyset$  & $\emptyset$ & 0 \\
\hline
\multicolumn{7}{l}{Output after iteration 3}\\
\hline
Sequence: & \sout{$x_1$} & \sout{$x_2$} & \sout{$(x_1, (x_2,x_3))$} & \sout{$x_4$} & \sout{$x_5$} & $((x_1, (x_2,x_3)),(x_4, (x_5,x_6)))$\\
Existential Probability: & 0 & 0 & 0 & 0 & 0 & 1\\
\hline
\end{tabular}
% \vspace{+.5em}
\caption{Simulation of an RvNN based on our new formulation. We use a strikethrough to illustrate ``non-existing" representations ($0$ existential probability) as they are ignored for further computation. We use $\emptyset$ for composition scores of non-existing representations. In each iteration, we bold the child with a composition probability of 1. We also use a $\rightarrow$ symbol to indicate that the child will be sent over to the first right position with existential probability of $1$ so that a composed parent representation can be formed at that position. The composition probabilities are predicted in each step by the decision function $D$.}
\vspace{-1mm}
\label{table:RvNNReformulation}
\end{table*}

\vspace{2mm}
\subsubsection{Decision function}
\vspace{2mm}
\setlength{\parskip}{0.5em}
\noindent {\bf Definition 2.} 
We define a decision function $D$ as a function dedicated to make structural decisions about which positions to compose together into a parent representation given the model state.

This definition generalizes both vanilla RvNNs and architectures such as Gumbel-Tree LSTMs. In case of vanilla RvNNs, $D$ can be conceived of as a trivial algorithm which makes the appropriate decision at every step by simply looking at the ground truth. In case of Gumbel-Tree LSTM, the function $D$ can represent its scoring function that scores all candidate representations to be chosen for composition in a particular step. In our generalized formulation, %we consider the decision function $D$ more generally. Here, 
$D$ makes structural decision by assigning a \textbf{composition probability} ($c_i$) to every position $i$ in a sequence $r_{1:n}$. This composition probability determines which positions are to be replaced by the parent representation. For now, we only consider $c_i \in \{0,1\}$. Note, unlike Gumbel-Tree LSTM, in our formulation multiple positions can be chosen for composition. In $\S$\ref{recursiveupdate}, we discuss the exact recursive rules to update the representations based on the composition probabilities. 

\subsubsection{Left and Right Functions}

\noindent {\bf Definition 3.} Given a sequence of values $v_{1:n} = (v_1, v_2,\cdots, v_n)$, we define a function $left$ which takes as input some value $v_i$ and returns the closest existing left value $v_j$ (i.e., $v_j = left(v_i)$) such that $e_j = 1$ and $\forall{l}, j<l<i,$ $\mbox{ } e_l = 0$. Note that since position $i-1$ can have an existential probability $e_{i-1}=0$, $left(v_i)$ is not always $v_{i-1}$.

In a similar vein, we also define a function $right$ 
%consider a function $right$ 
which takes as input some value $v_i$ and returns the closest existing  right value $v_j$ (i.e., $v_j = right(v_i)$) such that $e_j = 1$ and $\forall{l}, i<l<j,$ $\mbox{ } e_l = 0$. 
Similar to the $left$ function, $right(v_i)$ is not always $v_{i+1}$. Note that the sequence $v_{1:n}$ could be any sequence including a sequence of representations ($r_{1:n}$) or a sequence of composition probabilities ($c_{1:n}$).
\subsubsection{Recursive Update}
\label{recursiveupdate}

Now, we describe how the above components come together in each recursive update of our reformulated RvNN. We can formally express the rule for recursively updating some representation at position $i$, $r_i$ in the recursive step $k$ as: 
\begin{equation}
r^{k+1}_i = left(c_i) \cdot f (left(r_i^{k}), r_{i}^{k}) + (1-left(c_i)) \cdot r^{k}_i, \label{c_update}
\end{equation}
where $f$ refers to the composition function (recursive cell) as before and $c_i$ is the composition probability (as predicted by the function $D$) at any position $i$. Note that after this point, if $left(c_i)=1$, $left(r^{k}_i)$ is no longer needed because it has been composed together with $r^{k}_i$ and the composed parent representation already exists in position $i$ (as $r^{k+1}_i$) according to the above rule. Thus, when $left(c_i)=c_j=1$ we can set the existential probability ($e_j$) at position $left(i) = j$ as $0$. Thus, we can express the update rule for $e_j$ as:
\begin{equation}
{e}^{k+1}_j = {e}^{k}_j \cdot (1 - c_j). \label{e_update}
\end{equation}
In Table \ref{table:RvNNReformulation}, we show how we can simulate an RvNN based on the rules as laid out above using a sequence $x_{1:6}$.\footnote{Note, in iteration $2$ in Table \ref{table:RvNNReformulation}, when we compose together the representation of $x_1$ (position $1$) and the representation of $(x_2,x_3)$ (position $3$), the former is not at the immediate left (position $2$) from the latter because position $2$ has $0$ existential probability.}

\subsection{Towards Continuous RvNN}
In the above sections we  consider only discrete binary values for composition probability and existential probability. As such, it is still a ``hard RvNN" with a discrete order of composition. To transform it into a Continuous RvNN (soft RvNN), we simply incorporate a continuous relaxation to the decision function $D$ allowing it to predict composition probabilities ($c_i$) in the interval $[0,1]$. Similarly, we also allow existential probabilities ($e_i$) to be in $[0,1]$.

As a result of this relaxation, we can simply use a neural network with sigmoid activation for $D$. Given that we accommodate for continuous values, we do not have to force binary decisions using some form of reparameterization or gradient approximation. Moreover, we can directly use the recursive update rules as defined in Eq. \ref{c_update} and Eq. \ref{e_update} without any changes because they are already compatible with real-valued $c_i$ and $e_i$.

Below, we discuss our re-definition of $left$ and $right$ functions (neighbor retriever functions) under this new context.

\subsubsection{Neighbor Retriever Functions}
In our reformulation presented in \S\ref{reformulation}, an active role is played by the $left$ and $right$ functions. However, the previous definition (Def. 3) was made under the condition that the existential probabilities can only take binary values. Since now we have real-valued probabilities, the notion of ``closest existing left (or right) value" is not well defined because existential probabilities can have non-binary values.

Precisely, both $left$ and $right$ functions are important for CRvNN, and either of the functions can be re-purposed as the other with minimal changes. Thus, for now, we only focus on the $right$ function which can be easily adapted into a $left$ function.

\noindent {\bf Definition 4.} Given a sequence of values $v_{1:n}=(v_1,v_2,\cdots,v_n)$, and for every $v_i$, given a sequence of probabilities $p_{i1:in} = (p_{i1}, p_{i2},\cdots, p_{in})$ such that $p_{ij}$ indicates the probability that $v_j$ is the closest existing value right to $v_i$, we define the function $right$ as: $right(v_i) = \sum_{j=1}^n p_{ij} \cdot v_j$. 

Essentially, the function $right$ returns something analogous to an expected value of the immediate right existing representation. Note, however, that $\sum_j p_{ij}$ is not necessarily $1$, rather $\leq 1$. This is because there is another possibility that there is no existing representation at the right at all.

In our implementation, we transform the existential probabilities $e_{1:n}=(e_1,e_2,\cdots,e_n)$ into the sequence $p_{i1:in}$. Below, we formulate the precise 
rules that we use:
\begin{equation}
\label{adhoc-engineering-eqn}
p_{ij} =
\begin{cases}
    0 & j \leq i\\
    {e}_j & \sum_{l=i+1}^j {e}_l \leq 1\\
    \max(0,1-\sum_{l=i+1}^{j-1} {e}_l) & \sum_{l=i+1}^{j} {e}_l > 1\\
\end{cases}
\end{equation}

\begin{algorithm}[t]
   \caption{Continuous Recursive Neural Network}
   \label{alg:CRvNN}
\begin{algorithmic}
   \STATE {\bfseries Input:} data $x_{1:n} = (x_1, x_2,\cdots,x_n)$
   \STATE $r^1_i \gets leafTransform(x_i)$ \hfill\COMMENT{$i = 1, \cdots,n$}\\
   \color{black}
   \STATE Initialize $e^1_i \gets 1$ 
   \FOR{$k=1$ {\bfseries to} $n-1$}
   \STATE $c_{1:n} \gets decisionFunction(r^k_{1:n})$
   \STATE $\alpha \gets left(c_i)$
   \STATE $h \gets left(r^k_i)$
   \STATE $r^{k+1}_i \gets \alpha \cdot compose(h, r^k_i) + (1-\alpha) \cdot r^k_i$
   
   \STATE $e^{k+1}_i \gets e^k_i \cdot (1-c_i)$
   \IF{$dynamicHalt(e^{k+1}_{1:n})$}
   \STATE break
   \ENDIF
   \ENDFOR

\end{algorithmic}
\end{algorithm}

That is, we re-purpose the existential probabilities ${e}_j$ as $p_{ij}$. Naturally, we first set all non-right values from $i$ to have $0$ probability. Then we adjust them so that they sum up to $\leq 1$ by zeroing out $p_{ij}$ for any values after the position where the accumulated probability exceeds $1$.

To an extent, however, Eq. \ref{adhoc-engineering-eqn} is an engineering choice rather than being completely mathematically grounded. A more mathematically grounded choice would be to set $p_{ij} = e_j \cdot \Pi_{l=i+1}^{j-1} (1-e_l)$. We created a log-sum-exp based formulation for the equation (to avoid potential instabilities due to cumulative multiplication) as: 
\begin{equation}
  \label{math-grounded-eqn}
   p_{ij} = e_j \cdot \exp\left(\sum_{l=i+1}^{j-1} \log(1-e_l)\right)
\end{equation}
We did some experiments with this on the synthetic logical inference \cite{bowman-2015-tree} dataset. It performed reasonably but slightly worse than when we use Eq. \ref{adhoc-engineering-eqn}. Overall both formulations (Eq. \ref{adhoc-engineering-eqn} or Eq. \ref{math-grounded-eqn}) should be equivalent in the discrete setting, and thus, both can push the model to approximate a discrete RvNN. Regardless, for focus, we only consider Eq. \ref{adhoc-engineering-eqn} in the rest of the paper.

Given these formulations we can further generalize both $left$ and $right$ functions.

\noindent {\bf Definition 5.} Given a sequence $v_{1:n} = (v_1, v_2,\cdots, v_n)$, we define $left_m(v_i)$ as the function to retrieve the expected value in the $m^{th}$ left position from $i$. Formally:
\begin{equation}
    left_1(r_i) = left(r_i),
\end{equation}
\begin{equation}
    left_m(r_i) = left(left_{m-1}(r_i)).
\end{equation}
Similarly, we can define $right_m$. In the following subsection, we describe the main algorithm. 

\begin{algorithm}[t]
   \caption{\em dynamicHalt}
   \label{alg:dynamicHalt}
\begin{algorithmic}
   \STATE {\bfseries Input:} data $e_{1:n} = (e_1, e_2,\cdots,e_n)$ 
   \IF{$e_i < \epsilon$}  
   \STATE $d_i=0$ \hfill\COMMENT{$i = 1, \cdots,n$}\\
   \ELSE
   \STATE $d_i=1$
   \ENDIF
   \IF{$\sum_{i=1}^n d_i = 1$}
   \STATE return True
   \ELSE
   \STATE return False
   \ENDIF
\end{algorithmic}
\end{algorithm}

\subsection{Continuous RvNN: Algorithm}
The Continuous RvNN model is presented in Algorithm \ref{alg:CRvNN}. We already explained the recursive update rules. Next, we describe the implementation details of all the functions in the algorithm. 

\subsubsection{Leaf Transformation}
We use the function $leafTransform$ simply for an initial transformation of the embedded sequence. This can be expressed as:
\begin{equation}
    r_i^1 = LN(Wx_i + b).
\end{equation}
Here, $x_i \in {\rm I\!R}^{d_{embed} \times 1}$, $W \in {\rm I\!R}^{d_h \times d_{embed}}$, and $b \in{\rm I\!R}^{d_{h} \times 1} $. $LN$ refers to layer normalization \cite{ba2016layer}.

\subsubsection{Decision Function}
The decision function is used to predict the composition probabilities $c_i$. For this, we take into account the local context using a convolutional layer. However, we want the ``true local context" for which we need to use the $left_m$ and $right_m$ functions, as defined above. Given a locality window size of $2 \cdot \tau + 1$, we use the following function to get the initial un-normalized scores $u_i$ for computing the composition probabilities $c_i$:
\begin{equation}
    u_i = W_2\mathrm{GeLU}\left(\sum_{j=-\tau}^{\tau} W_{conv}^{j+\tau}L(r_i, j) + b_{conv}\right) + b_2,
\end{equation}
\begin{equation}
L(r_i, j) =
\begin{cases}
    left_{-j}(r_i) & j < 0,\\
    r_i & j = 0,\\
    right_{j}(r_i) & j > 0.\\
\end{cases}
\end{equation}
Here, we have a set of convolutional kernel weights $\{W_{conv}^0, W_{conv}^1, , \cdots, W_{conv}^{2\cdot \tau}\}$, where any $W_{conv}^l \in {\rm I\!R}^{d_h \times d_h}$; $L(r_i, j) \in {\rm I\!R}^{d_h \times 1}$, $W_2 \in {\rm I\!R}^{1 \times d_h}$, $b_{conv} \in {\rm I\!R}^{d_h}$ and, $b_2 \in {\rm I\!R}^{1}$. Note that $u_i$ is now an un-normalized scalar real value. To turn it into a probability we can use $sigmoid$:
\begin{equation}
    c_i = sigmoid(u_i) = \frac{exp(u_i)}{exp(u_i)+1}.
\end{equation}
The use of sigmoid allows multiple positions to have high composition probabilities ($c_i$). As such, multiple parents can be composed in the same recursion resulting in the parallelism that we alluded to before. However, pure sigmoid in this form is very unconstrained. That is, there is no constraint that prevents multiple contiguous (as defined by $left$ and $right$ functions) positions from having high composition probabilities ($c_i$). Concurrently composing contiguous representations with their expected right representation will violate the tree-structure. It can also cause information being propagated to positions that, at the same time, lose their own existential probability due to propagating their information rightwards. As such, some propagated information can henceforth be ignored due to landing in an area that loses its existential probability.
 Thus, to prevent contiguous positions from having high existential probabilities, we \textbf{modulate} the sigmoid formulation with the scores from the neighbors as follows: 
\begin{equation}
    c_i = \frac{exp(u_i)}{exp(u_i) + exp(left(u_i)) + exp(right(u_i)) +1}.
\end{equation}
We refer to this function as \textbf{modulated sigmoid}.

\subsubsection{Composition Function}
\label{composition}
We use the same recursive gated cell as introduced by \citet{shen-2019-ordered} for our composition function. This is originally inspired from the feedforward functions of Transformers \cite{vaswani-etal-2017-attention}.
\begin{equation}
    \left[
    \begin{matrix}
        z_i \\ 
        h_i \\
        c_i \\
        u_i
    \end{matrix}
    \right]=  W_2~ \mathrm{GeLU} \left( W^{Cell}_1  \left[ 
    \begin{matrix}
        left(r_i) \\ 
        r_i
    \end{matrix} 
    \right] 
    + b_1 \right) + b_2
\end{equation}
\begin{eqnarray}
o_i = LN (\sigma(z_i) \odot left(r_i)
                    + \sigma(h_i) \odot r_i 
                    + \sigma(c_i) \odot u_i) \label{eq:gated_sum}
\end{eqnarray}
Here, $\sigma$ is $sigmoid$; $o_i$ is the output parent $\in {\rm I\!R}^{d_h \times 1}$; $r_i, left(r_i) \in {\rm I\!R}^{d_h \times 1}$;
$W_1^{cell} \in {\rm I\!R}^{d_{cell} \times 2\cdot d_h}$; $b_1 \in {\rm I\!R}^{d_{cell} \times 1}$; $W_2 \in {\rm I\!R}^{d_{h} \times d_{cell}}$; $b_1 \in {\rm I\!R}^{d_{h} \times 1}$.
Different from \cite{shen-2019-ordered}, we use GeLU \cite{DBLP:journals/corr/HendrycksG16} instead of ReLU as the activation function.

\begin{table*}[t]
\small
\centering
\def\arraystretch{1.2}
\begin{tabular}{  l | l l l l l l | l l l} 
%\toprule
\hline
\textbf{Model} & \multicolumn{6}{c}{\textbf{Number of Operations}} & \multicolumn{3}{|l}{\textbf{Systematicity}}\\
& 7 & 8 & 9 & 10 & 11 & 12 & A & B & C\\
\hline

%\midrule
\multicolumn{7}{l}{(\textit{Sentence representation models + ground truths})}\\
\hline
Tree-LSTM* & $94$ & $92$ & $92$ & $88$ & $87$ & $86$ & $91$ & $84$ & $76$\\
Tree-Cell* & $98$ & $96$ & $96$ & $95$ & $93$ & $92$ & $95$ & $95$ & $90$\\
Tree-RNN* & $98$ & $98$ & $97$ & $96$ & $95$ & $96$ & $94$ & $92$ & $86$\\
\hline
\multicolumn{7}{l}{(\textit{Inter-sentence interaction models})}\\
\hline
Transformer* & $51$ & $52$ & $51$ & $51$ & $51$ & $48$ & $53$ & $51$ & $51$\\
Universal Transformer* & $51$ & $52$ & $51$ & $51$ & $51$ & $48$ & $53$ & $51$ & $51$\\
\hline
\multicolumn{7}{l}{(\textit{Sentence representation models})}\\
\hline
LSTM* & $88$ & $84$ & $80$ & $78$ & $71$ & $69$ & $84$ & $60$ & $59$\\
RRNet* & $84$ & $81$ & $78$ & $74$ & $72$ & $71$ & --- & --- & ---\\
ON-LSTM* & $91$ & $87$ & $85$ & $81$ & $78$ & $75$ & $70$ & $63$ & $60$\\
Ordered Memory* & $98_0$ & $97_4$ & $96_5$ & $94_8$ & $93_5$ & $92_{11}$ & $94$ & $91$ & $81$\\
\hline
\multicolumn{7}{l}{(\textit{Our model})}\\
\hline
CRvNN & $98_1$ & $97_3$ & $96_2$ & $95_6$ & $94_8$ & $93_5$ & $99_1$ & $98_3$ & $92_{22}$\\
\hline
\end{tabular}
%\vspace{+.5em}
\caption{Accuracy on the Synthetic Logical Inference dataset \cite{bowman-2015-tree} for different number of operations after training on samples with $\leq 6$ operations. We also show results of testing systematicity of the models in specially designed splits: A, B, and C. * indicates that the results were taken from \cite{shen-2019-ordered}. Our models were run $5$ times on different random seeds. We show its mean and standard deviation. Specifically, subscript represents standard deviation. As an example, $90_1 = 90\pm0.1$}.
\label{table:PNLIRESULTS}
\vspace{-4mm}
\end{table*}

\subsubsection{Dynamic Halt}
Ideally, CRvNN can learn to simultaneously process all children at the same hierarchy level. Thus, CRvNN will only need to recurse over the tree depth. This means that we need a mechanism to detect if the induced tree-depth has been traversed and if we can halt. Following our framework, we only need to look at the sequence of existential probabilities $e_{1:n} = (e_1,e_2, \cdots, e_n)$. Near the ideal time for halting, all existential probabilities except the last should be close to $0$ ($0$ indicates that it is no longer in need of further processing). The last position has nothing to the right of it to be composed. So we enforce the last position to always have $0$ composition probability and thus, $1$ existential probability. Overall, to determine when to halt, we simply check if all existential probabilities except the last are less than some small threshold $\epsilon$. Algorithm \ref{alg:dynamicHalt} shows an implementation of this mechanism.
\subsection{Extra components}
Here we describe some extra, potentially less essential, components that we use in our implementation of CRvNN. 
\subsubsection{Special Embeddings}
We prepend all sequences with a special $<$START$>$ embedding, and we append all sequences with a special $<$END$>$ embedding. Both are trainable vectors of size $d_{embed}$. These embeddings are enforced to always have a composition probability $0$ and an existential probability of $1$. When constructing the local context, these embeddings can provide more explicit information about how close to the starting or ending boundary a representation $r_i$ is. 
\subsubsection{Transition features}
\label{transition}
In an attempt to enhance the decision function, we also concatenate features to provide the explicit information if a position $i$ was composed and updated in the last recursion or not. For that, we construct a set of features $T \in {\rm I\!R}^{d_{s}}$ as:
\begin{equation}
    T = left(c_i) \cdot C + (1-left(c_i)) \cdot \widehat{C}.
\end{equation}
Here, $C$ and $\widehat{C}$ are both trainable vector parameters $\in {\rm I\!R}^{d_{s}}$. $C$ can be conceived to be representing a set a features indicating that in the last iteration the concerned position was composed with the left child and updated, whereas $\widehat{C}$ can be conceived to be representing the contrary. Here, we use the $left(c_i)$ values from the last recursion (initially $0$).

\subsubsection{Halt Penalty}
\label{haltpenalty}
As discussed before, during halting, in the sequence of existential probabilities ($e_{1:n}$) all but the last existential probabilities should be close to $0$. While this is ideally supposed to happen if all the positions are processed properly, there is no guarantee that it will indeed happen. To encourage this property, we use an auxiliary loss ($A(e_{1:n})$), which we define as:
\begin{equation}
    A(e) = -\log\left(\frac{e_n}{\sum_{j=1}^n e_j}\right).
\end{equation}
This can be conceived as cross entropy between the ideal final sequence of existential probabilities (all but the last being $0$) and the actual final sequence of the same after normalization.
The overall optimization function can be formulated as:
\begin{equation}
\vspace{-3mm}
\min_{\theta} \mathcal{L}(\theta) + \gamma\cdot A(e),
\end{equation}
where $\mathcal{L}(\theta)$ is the main cross entropy loss and $\gamma$ is a Lagrangian multiplier for the auxiliary objective.

\vspace{-2mm}
\section{Experiments and Results}
In this section, we discuss our experiments and results. We evaluate our model on logical inference \cite{bowman-2015-tree}, list operations (ListOps) \cite{nangia-bowman-2018-listops},  sentiment analysis---two datasets, SST2 and SST5 \cite{socher-etal-2013-recursive}, and natural language inference---two datasets, SNLI \cite{snli:emnlp2015} and MNLI \cite{N18-1101}. For implementation details, refer to the appendix. 

\begin{table*}[t]
\small
\centering
\def\arraystretch{1.2}
\begin{tabular}{ l | r | r | r  | r | r | r | r | r} 
%\toprule
\hline
 Model & \multicolumn{8}{c}{\textbf{Sequence length ranges (ListOps)}} \\
 & $200-300$ & $300-400$ & $400-500$ & $500-600$ & $600-700$ & $700-800$ & $800-900$ & $900-1000$ \\
\hline
CRvNN & $98.51\pm1.1$ & $98.46\pm1.3$ & $98.04\pm 1.3$ & $97.95\pm 1.1$ & $97.17\pm 1.6$ & $97.84\pm 1.7$ & $96.94\pm 1.6$ & $96.78\pm 1.9$\\
\hline
\end{tabular}
\caption{Extrapolation of CRvNN on ListOps after training on samples of length $\leq$ 100. We used the ListOps extrapolation test set from \cite{havrylov-etal-2019-cooperative}.
 } 
\label{table:listops_extrapolation}
\end{table*}

\subsection{Logical Inference}
In the logical inference dataset, we focus on two particular generalization properties separately - length generalization and compositional generalization (i.e., systematicity).
We compare
CRvNN with 
Tree-LSTM \cite{tai-etal-2015-improved}, Tree-Cell \cite{shen-2019-ordered}
Tree-RNN \cite{bowman-2015-tree}, Tranformer \cite{vaswani-etal-2017-attention}, Universal Transformer \cite{dehghani-etal-2018-universal},
LSTM \cite{hochreiter-etal-1997-long}, RRNet \cite{jacob-etal-2018-learning}, ON-LSTM \cite{shen-2018-ordered}, Ordered Memory \cite{shen-2019-ordered} (see Table \ref{table:PNLIRESULTS}). 

\begin{table} [t]
\small
\centering
\def\arraystretch{1.2}
\begin{tabular}{  l |  l} 
%\toprule
\hline
\textbf{Model} & \textbf{Accuracy}\\
\hline

%\midrule
\multicolumn{2}{l}{(\textit{Models with ground truth})}\\
\hline
Tree-LSTM$\ddagger$  & $98.7$ \\
\hline
\multicolumn{2}{l}{(\textit{Models without ground truth})}\\
\hline
Transformer* & $57.4\pm0.4$ \\
Universal Transformer* {\color{white}xxx} & $71.5\pm7.8$ \\
LSTM $\dagger$ & $71.5\pm1.5$ \\
RL-SPINN $\dagger$ & $60.7\pm2.6$\\
Gumbel-Tree LSTM $\dagger$ & $57.6\pm2.9$\\
\cite{havrylov-etal-2019-cooperative} $\dagger$ & $99.2 \pm 0.5$\\
Ordered Memory* & $99.97\pm0.014$ \\
\hline
\multicolumn{2}{l}{(\textit{Our model})}\\
\hline
CRvNN & $99.6\pm0.3$ \\
\hline

\hline
\end{tabular}
\caption{Accuracy on ListOps. Results with * were taken from \cite{shen-2019-ordered}. $\ddagger$ indicates that the results were taken from \cite{nangia-bowman-2018-listops}. $\dagger$ indicates that the results were taken from \cite{havrylov-etal-2019-cooperative}. Our models were run $5$ times on different random seeds. We show its mean and standard deviation.} 
\vspace{-7.5mm}
\label{table:LISTOPSRESULTS}
\end{table}

\subsubsection{Length Generalization}
To evaluate CRvNN for length generalization, as in prior work, we train the model only on samples with $\leq 6$ operations whereas we test it on samples with higher unseen number of operations ($\geq 7$). In Table \ref{table:PNLIRESULTS}, discounting the performance of RvNN-based models with ground-truth access, our model, along with ordered memory, achieves the best performance on length generalization.

\begin{table}[t]
\small
\centering
\def\arraystretch{1.2}
\begin{tabular}{  l | l | l | l | l} 
%\toprule
\hline
\textbf{Model} & \textbf{SST2} & \textbf{SST5} & \textbf{SNLI} & \textbf{MNLI}\\
\hline
RL-SPINN$\ddagger$ & --- & --- & $82.3$ & $67.4$ \\
Gumbel-Tree-LSTM$\dagger\dagger$ & $90.7$ & $53.7$ & $85.6$ & --- \\
Gumbel-Tree-LSTM$\ddagger$ & --- & --- & $83.7$ & $69.5$ \\
Gumbel-Tree-LSTM$\dagger$ & $90.3_5$ & $51.6_8$ & $84.9_1$ & --- \\
\cite{havrylov-etal-2019-cooperative}$\dagger$ & $90.2_2$ & $51.5_4$ & $85.1_2$ & $70.7_3$ \\
Ordered Memory* & $90.4$ & $52.2$ & --- & --- \\
%Ordered Memory & --- & --- & $85.2_1$ & 72.8_3 \\
CRvNN & $88.3_6$ & $51.4_{13}$ & $85.1_2$ & $72.9_2$ \\
\hline
\end{tabular}
%\vspace{+.5em}
\caption{Accuracy on multiple natural language datasets. * indicates that the results were taken from \cite{shen-2019-ordered}. $\dagger$ indicates that the results were taken from \cite{havrylov-etal-2019-cooperative}.  $\ddagger$ indicates that the results were taken from \cite{TACL1281}. $\dagger\dagger$ indicates that the results were taken from \cite{choi-2018-learning}. Our models were run $5$ times (except, on MNLI it was run $3$ times) on different random seeds. We show its mean and standard deviation. Subscript represents standard deviation. E.g., $90_1 = 90\pm0.1$.} 
\label{table:natural}
\end{table}

\subsubsection{Systematicity}
Following \citet{shen-2019-ordered}, we create three different train-test splits on the logical inference dataset: A, B, and C (with increased level of difficulty from A to C, A being the easiest). For each split, we filter all samples with a specific compositional pattern from the training set and put them into the test set. Then we check whether our model can generalize to unseen patterns (unseen combinations of operands and operators). In set A, we filter samples according to the pattern $*( and (not\,a) ) *$, in B, we filter according to the pattern $* ( and (not *) ) *$, and in C, we filter according to $* ( \{and,or\} (not *) ) * $. 
As evident in Table \ref{table:PNLIRESULTS}, CRvNN exhibits exceptional capability for compositional generalization and outperforms 
all prior reported results. 

\subsection{ListOps}
ListOps is a challenging synthetic task which explicitly demands capability for hierarchical modeling. Several prior latent-tree models were shown to perform poorly on it \cite{nangia-bowman-2018-listops}. As shown in Table \ref{table:LISTOPSRESULTS}, CRvNN gets close to perfect accuracy, demonstrating its capability to capture underlying structures without structural supervision. We also show the length generalization capability of CRvNN on ListOps in Table \ref{table:listops_extrapolation}. CRvNN still achieves high accuracy in much higher sequence lengths ($400-1000$) even when it is originally trained on sequences of lengths $\leq 100$.

\subsection{Natural Language Datasets}
In Table \ref{table:natural}, we also evaluate CRvNN on natural language datasets (SST2, SST5, SNLI, MNLI). Consistent with prior work \cite{havrylov-etal-2019-cooperative}, for MNLI we augment the training data with SNLI training data and report the test result on the matched test set. In the real-world tasks, CRvNN obtains mixed results, but overall the results are comparable to prior work in the similar latent-tree context. Particularly, CRvNN performs comparatively weak on SST2, but on contrary it performs significantly better than prior work (besides OM) on MNLI, which is a harder task. We also ran a CRvNN model with an enforced balanced tree structure induction on MNLI but its performance was comparatively worse ($71.5 \pm 0.4$). For Gumbel-Tree LSTM \citet{choi-2018-learning}, we show results reported from different works because the results fluctuate from one paper to another \cite{havrylov-etal-2019-cooperative, TACL1281}. 

\begin{table*}[t]
\small
\centering
\def\arraystretch{1.2}
\begin{tabular}{  l | r  | r | r | r | r} 
%\toprule
\hline
\textbf{Model} & \multicolumn{5}{c}{\textbf{Sequence length ranges}} \\
& $81-100$ & $101-200$ & $201-500$ & $501-700$ & $701-1000$ \\
\hline
Ordered Memory & 3.28 min & 5.54 min & 13.23 min & 25 min & 35.34 min\\
CRvNN & 0.20 min & 0.52 min & 0.30 min & 0.38 min & 1.08 min\\
\hline

\hline
\end{tabular}
%\vspace{+.5em}
\caption{Training time taken by Ordered Memory and CRvNN models to be run on $50$ samples on various sequence length ranges. We used the publicly available code to run Ordered Memory.} 
\label{table:stresstest}
% \vspace{-3mm}
\end{table*}
\begin{table}[t]
\small
\centering
\def\arraystretch{1.2}
\begin{tabular}{  l | l l l  | l} 
%\toprule
\hline
\textbf{Model} & \multicolumn{3}{c}{\textbf{Logical inference}} & \multicolumn{1}{|c}{\textbf{ListOps}}\\
& \multicolumn{3}{c|}{Number of Operations} \\
& 10 & 11 & 12 & \\
\hline
CRvNN & $95.0_6$ & $93.7_8$ & $93.1_5$ & $99.6_3$ \\
 -- modulated sigmoid & $95.0_{3}$ & $94.3_4$ & $92.8_5$ & $90.8_{163}$\\
 -- transition features & $95.0_8$ & $94.4_7$ & $92.1_5$ & $99.4_5$ \\
 -- structure & $87.2_{11}$ & $86.1_{10}$ & $80.9_{20}$ & $83.4_{11}$\\
 -- halt penalty & $95.2_5$ & $94.6_4$ & $93.1_7$ & $99.4_2$\\
 -- GeLU + ReLU & $95.1_{7}$ & $94.1_{5}$ & $92.7_{6}$ & $99.2_4$ \\
 -- Cell + LSTMCell & $89.4_{7}$ & $89.2_{7}$ & $86.5_{11}$ & $71.2_{41}$\\
\hline
\end{tabular}
\caption{Cell represents Gated Recursive Cell. All the models were run $5$ times on different random seeds. Subscript represents standard deviation. For example, $90_1 = 90\pm0.1$} 
\label{table:ablation}
\end{table}
\begin{table}[t]
\small
\centering
\def\arraystretch{1.2}
\begin{tabular}{  l } 
%\toprule
\hline
\textbf{Parsing Examples}\\
\hline
((((i did) not) like) (((a single) minute) ((of this) (film .))))\\
\hline
(((roger dodger) ((is (one of)) (the most)))\\ (compelling (variations (of (this theme)))))\\
\hline
\end{tabular}
\caption{Parsing examples obtained using CRvNN trained on MNLI. The example sentences are from \cite{socher-etal-2013-recursive}.} 
\label{table:parsetrees}
\vspace{-3mm}
\end{table}
\section{Analysis}
In this section, we contrast CRvNN with  Ordered Memory in terms of speed and show an ablation study of CRvNN. 
\subsection{Speed Test}
\label{stresstest}
Ordered memory (OM) is a close competitor for CRvNN. Both of the models are end-to-end differentiable. Both can simulate an RvNN without ground truth structures. CRvNN neither consistently nor conclusively outperforms OM. However, there is one crucial advantage for CRvNN. Particularly, it can achieve a degree of parallelism by processing multiple positions concurrently. At the same time, it can do an early-exit using dynamic halt. In contrast, OM not only has an outer loop over the whole sequence length, but it also has an inner loop where it recurse over its memory slots, adding significant overhead. The inner loop recursion of OM involves heavy sequential matrix operations. As such, in practice, CRvNN can be much faster than OM. In Table \ref{table:stresstest}, we compare the training run time for both models. We generated synthetic ListOps samples for different ranges of sequence lengths. For each range of sequence lengths, we trained both the models on $50$ samples for $1$ epoch and $1$ batch size on an AWS $P3.2\times$ instance (Nvidia V$100$). As we can see from the table, CRvNN is substantially faster than OM. Although in other settings with different tasks and different batch sizes, the gap between the speed of OM and CRvNN may not be as high as in Table \ref{table:stresstest}, we still noticed CRvNN to be roughly $2-4$ times faster than OM even when using higher batch sizes for OM (taking advantage of its lower memory complexity). We also tried running CYK-LSTM \cite{maillard_clark_yogatama_2019} but faced memory issues when running it for longer sequences ($>200$ length). 

\subsection{Ablation Study}
In Table \ref{table:ablation}, we show an ablation study on CRvNN. We find that replacing modulated sigmoid with simple sigmoid significantly degrades and destabilizes the performance in ListOps. Removing ``structure" or the structural bias by simply using the composition function as left-to-right recurrent network, again, significantly harms the performance of the model. The gated recursive cell itself is also crucial for the performance. Replacing it with an LSTM cell \cite{hochreiter-etal-1997-long} causes severe degradation. To an extent, this is consistent with the findings of \citet{shen-2019-ordered}. They replaced the gated recursive cell in ordered memory \cite{shen-2019-ordered} with an RNN cell, and observed substantial degradation. However, even with an LSTM cell, CRvNN performs better, in logical inference, than any of the other reported models which do not have the gated recursive cell. Replacing the activation function, removing the transition features (\S\ref{transition}), or eliminating the halt penalty (\S\ref{haltpenalty}) makes little difference.

\subsection{Parsing Results}
The internal composition scores of each layer of CRvNN can be used to parse induced trees. While there can be multiple ways to convert the composition scores to extract trees, one method is to simply treat any particular position of a given sequence at a particular iteration in the algorithm as having a composition probability of $1$ whenever the cumulative composition probability at that position over all the iterations thus far is $\geq 0.5$. Otherwise, we treat the position at the particular iteration as having a composition probability of $0$. Once the scores are binarized, it is straight-forward to extract the trees following the ideas discussed in  \S\ref{reformulation}. Table \ref{table:RvNNReformulation} (in \S\ref{reformulation}) shows an example on how binary and discrete composition probabilities relate to a particular structure of composition. In Table \ref{table:parsetrees}, we show parsing examples obtained with CRvNN using the above procedure. 

\section{Related Work}
\label{Related Work}
Several early approaches focused on adapting neural models for simulating pushdown automata or for grammar induction in general \cite{sun-1990-ijcnn-connectionist, giles-1990-nips-higher, das-1992-cogsci-learning, mozer-1993-nips-a, zeng-1994-ieee-discrete, grefenstette-2015-nips}. Also, recently, there are multiple works that focus on structure induction based on language modeling objectives %scicluna-de-la-higuera-2014-pcfg,
 \cite{yogatama-2018-iclr-memory, shen-2018-iclr-neural, shen-2018-ordered, li-etal-2019-imitation, kim-etal-2019-unsupervised, drozdov-etal-2019-unsupervised-latent, drozdov-etal-2020-unsupervised, shen2021structformer}.
In this work, we focus on models with structural bias that are used for general downstream tasks including NLP tasks such as classification and NLI. 

\citet{pollack-1990-recursive} presented RvNN as a recursive architecture to compose tree and list-like data-structures. \citet{Socher10learningcontinuous, socher-etal-2013-recursive} aimed at capturing syntactic and semantic patterns of linguistic phrases and showed the effectiveness of RvNNs for sentiment analysis. Originally, RvNNs relied on external parsers to provide tree-structured inputs. Several works focused on augmenting RvNNs so that they can automatically induce the tree structure from plain text. To this end, \citet{le-zuidema-2015-forest} used a chart-based parsing algorithm (Cocke–Younger–Kasami algorithm \cite{sakai1962syntax}) in a neural framework with a convolutional composition function. \citet{maillard_clark_yogatama_2019} extended it with an attention mechanism to pool over multiple subtree candidates and a Tree-LSTM \cite{tai-etal-2015-improved} composition function. \citet{bowman-etal-2016-fast} presented a stack-augmented RNN (SPINN) to simulate the functions of a Tree-RNN based on the principles of shift-reduce parsing \cite{schutzenberger-1963-ic-on, knuth-1965-ic-on} but relied on ground truth structure annotations for supervision. \citet{yogatama-2017-iclr-learning} augmented SPINN with reinforcement learning (RL-SPINN) allowing unsupervised structure induction. \citet{maillard-clark-2018-latent} enhanced shift-reduce parsing-based stack-RNNs with beam search. \citet{munkhdalai-yu-2017-neural} introduced Neural Tree Indexers and gained strong performance by using full binary trees on natural language tasks. Similarly, \citet{shi-etal-2018-tree} showed that trivial trees (e.g., binary balanced trees) can be used to get competitive performances on several natural language tasks. \citet{dong2018neural} proposed a neural model to learn boolean logic rules and quantifications to reason over relational data with multiple arities.
\citet{choi-2018-learning} used Gumbel Softmax \cite{jang-2017-} to adaptively make discrete decisions at every recursive step in choosing a parent to compose with a tree-LSTM. \citet{jacob-etal-2018-learning} proposed a combination of recursive and recurrent neural structure through reinforcement learning for hierarchical composition. \citet{havrylov-etal-2019-cooperative} extended Gumbel Tree-LSTM \cite{choi-2018-learning} through disjoint co-operative training of the parser (with reinforcement learning) and the  composition function. \citet{shen-2019-ordered} developed ordered memory (OM) by synthesizing the principles of ordered neurons \cite{shen-2018-ordered} with a stack-augmented RNN. While OM is a strong contender to CRvNN, we discussed the advantages of CRvNN as a more parallelized sequence processor in \S\ref{stresstest}.

In another direction, \citet{liu-lapata-2018-learning} applied structured attention \cite{kim2017} based on induced dependency trees to enhance downstream performance. \citet{pmlr-v80-niculae18a} introduced SparseMAP to apply marginalization over a sparsified set of latent structures whereas \citet{corro-titov-2019-learning} did a Monte-Carlo estimation by stochastically sampling a projective depedency tree structure for their encoder. These methods, similar to ours, are end-to-end differentiable, but do not focus on hierarchical composition of a sequence vector (root node) from the elements (leaf nodes) in the sequence following its latent constituency structure. As such, they  do not address the challenges we face here.

In recent years, Transformers \cite{vaswani-etal-2017-attention} have also been extended either to better support tree structured inputs \cite{shiv2019novel, ahmed-etal-2019-need} or to have a better inductive bias to induce hierarchical structures by constraining self-attention \cite{wang-etal-2019-tree, Nguyen2020Tree-Structured, shen2021structformer} or by pushing intermediate representations to have constituent information \cite{fei-etal-2020-retrofitting}. However, the fundamental capability of Transformers for composing sequences according to their latent structures in a length-generalizable manner is shown to be lacking \cite{tran-etal-2018-importance, shen-2019-ordered, hahn-etal-2020-theoretical}.

% \vspace{+0.6em}

\section{Conclusion and Future Directions}
\label{Conclusion}
We proposed a reformulation of RvNNs to allow for a continuous relaxation of its structure and order of composition. The result is CRvNN, which can dynamically induce structure within data in an end-to-end differentiable manner. One crucial difference from prior work is that it can parallely process multiple positions at each recursive step and also, dynamically halt its computation when needed. We evaluated CRvNN on six datasets, and obtain strong performance on most of them. There are, however, several limitations of the model. First, the neighbor retriever functions in CRvNN, construct an $n \times n$ matrix ($n$ is the sequence length) which can be memory intensive. This is similar to the memory limitations of a Transformer. Another limitation is that it is a greedy model. It is also not explicitly equipped to handle structural ambiguities in natural language. In future work, we will address these limitations. To mitigate the memory limitations, we plan to constrain the neighbor retriever functions to look at only $k$ left or right candidates so that we only need an $n \times k$ matrix. We can compress contiguous positions with low existential probabilities so that we do not need to look beyond $k$. To handle its greedy nature, we will extend CRvNN to follow multiple paths concurrently. 

% \vspace{-2mm}
\section{Acknowledgment}
We would like to sincerely thank our reviewers for their constructive feedback that helped us to improve our paper greatly. We also thank Adrian Silvescu for helpful discussions. 
This research is supported in part by NSF CAREER award \#1802358, NSF CRI award \#1823292, and an award from UIC Discovery Partners Institute.
The computation for this project was performed on Amazon Web Services.
\bibliography{main}
\bibliographystyle{icml2021}
\begin{table*}[th]
\small
\centering
\def\arraystretch{1.2}
\begin{tabular}{  l | r | r | r | r | r } 
%\toprule
\hline
\textbf{Dataset} & \thead{\textbf{Initial} \\\textbf{Embedding} \\\textbf{Size}\\ ($d_{initial\_embed}$)} & \thead{\textbf{Hidden Size}\\($d_h$ or $d_{embed}$)} & \thead{\textbf{Input}\\ \textbf{Dropout}} & \thead{\textbf{Output}\\ \textbf{Dropout}} & \thead{\textbf{Hidden}\\ \textbf{Dropout}} \\
\hline
Logical Infer. & 200 & 200 & 0.1 & 0.3 & 0.1\\
ListOps & 128 & 128 & 0.3 & 0.2 & 0.1\\
SST2 & 300 & 300 & 0.3 & 0.2 & 0.4\\
SST5 & 300 & 300 & 0.3 & 0.2 & 0.4\\
SNLI & 300 & 300 & 0.4 & 0.1 & 0.1\\
MNLI & 300 & 300 & 0.4 & 0.1 & 0.1\\
\hline
\end{tabular}
%\vspace{+.5em}
\caption{Hyperparameter details for CRvNN.} 
\label{table:CRvNNhp}
\end{table*}

\newpage 

\appendix
\section{Architecture Details}
For every task used in our experiments we use an initial affine transformation where the initial embeddings of size $d_{initial\_embed}$ are transformed into the size $d_{embed}$. Typically, we set $d_{embed}$ as $d_{h}$. See Table \ref{table:CRvNNhp} for their values.

We treat the last representation in the sequence after being processed by CRvNN as the sentence encoding constructed by CRvNN. 

For classification tasks, we classify the sentence encoding by transforming it into logits for the classes after passing it through a series of affine layers (typically, 1 or 2). Intermediate layers have $d_{h}$ neurons where $d_h$ is also the dimension of the sentence encoding. 

For inference tasks (requiring sequence-pair comparison), like logical inference or SNLI and MNLI, we use a Siamese framework. Concretely, we first encode (using the same encoder with same parameters) both the premise and hypothesis (separately) into sentence vectors, say,  $s_1$ and $s_2$ respectively (both with $d_h$ dimensions). Then, we construct a classification feature vector $o$ as:
\begin{equation}
    o = [s_1;s_2;|s_1-s_2|;s_1\odot s_2]
\end{equation}
Here, $[;]$ indicates concatenation. We send $o$ to  a Multi Layer Perceptron (MLP) to classify the sequence relationship. The final layer activation is $Softmax$, but if there are intermediate MLP layers, we use $GeLU$ for them. We use a dropout (hidden dropout) in the gated recursive cell (after its first affine transformation). We use a dropout (input dropout) on the input just before sending it to CRvNN. We use another dropout (output dropout) in between the final MLP layers. All our models were trained on AWS p$3.2\times$ instance (Nvidia v$100$).

\section{Implementation Details}
For all experiments, as an optimizer, we use Ranger \cite{Ranger}\footnote{{\url{https://github.com/lessw2020/}}} or Rectified \cite{Liu2020On} Adam \cite{adam} with lookahead ($k=5, \alpha=0.8$) \cite{lookahead} and decorrelated weight decay \cite{loshchilov2018decoupled} ($1e-2$) with a learning rate of $1e-3$. We used GloVe ($300$ dimensions, $840$B) \cite{pennington2014glove} as un-trainable embeddings for natural language data. We set the cell size ($d_{cell}$ as referred in \S\ref{composition}) as $4 \cdot d_h$ ($d_h$ is the hidden size). We set the size of transition features ($d_s$ as referred in \S\ref{transition}) as $64$. For convolution in the decision function, we always use a window size of $5$. For halt penalty in \S\ref{haltpenalty}, we set $\gamma$ as $0.01$. Generally, we use a two-layered MLP on the sentence encoding from CRvNN. However, on ListOps, we used a single-layer. We use a batch size of $128$ for all tasks. We describe other hyperparameters of CRvNN in Table \ref{table:CRvNNhp}. We cut the learning rate by half if the validation loss does not decrease for $3$ contiguous epochs. 

\vspace{1mm}
\section{Hyperparameter Search}
On ListOps, we tune different dropouts among $\{0.1, 0.2, 0.3, 0.4\}$ separately using grid search (we ran for $10$ epochs and $50,000$ subsamples). For the Logical Inference task (length generalization task), we tune the different dropouts among $\{0.1, 0.2, 0.3\}$ for $7$ epochs per trial using grid search. We use the same hyperparamters for systematicity splits. For SST5, we tune the dropouts in $\{0.2, 0.3, 0.4\}$ for $3$ epochs. For SNLI, we tune the different dropouts among $\{0.1, 0.2, 0.3, 0.4\}$ for $5$ epochs, using a sub-sample of $100K$ examples, and for a maximum of $20$ trials using Tree of Parzen Estimators (TPE) \cite{10.5555/2986459.2986743}. We use Hyperopt \cite{10.5555/3042817.3042832} for hyperparameter tuning. For other components we mostly use similar hyperparameters as \citet{shen-2019-ordered} or default settings. We share the hyperparameters found for SST5 with SST2 and we also share the hyperparameters found for SNLI with MNLI.

\vspace{1mm}
\section{Datasets}
For all datasets, we use the standard splits as used by prior work. For training efficiency, we filter out training samples of sequence size $>150$ from MNLI. We filter out training samples with sequence length $>100$ from ListOps. We use the $90K$ sample version of ListOps similar to prior work.

%%%%%%%%%%%%%%%%%%%%%%%%%%%%%%%%%%%%%%%%%%%%%%%%%%%%%%%%%%%%%%%%%%%%%%%%%%%%%%%
%%%%%%%%%%%%%%%%%%%%%%%%%%%%%%%%%%%%%%%%%%%%%%%%%%%%%%%%%%%%%%%%%%%%%%%%%%%%%%%

\end{document}